\documentclass[journal=jacsat,manuscript=article]{achemso}

\usepackage{graphicx}%
\usepackage{multirow}%
\usepackage{amsmath,amssymb,amsfonts}%
\usepackage{amsthm}%
\usepackage{mathrsfs}%
\usepackage{titlesec}
\usepackage{booktabs}%
\usepackage{algorithm}%
\usepackage{algpseudocode}%
\usepackage{listings}%
\usepackage{array}
\usepackage{hypdoc}
\usepackage{microtype}
\usepackage[font=footnotesize,labelfont=bf]{caption}
\usepackage{float}
\usepackage{natbib}
\usepackage{setspace}
\usepackage{xkeyval}
\usepackage{times}
\usepackage{makecell}
\usepackage{chemformula} 
\usepackage[T1]{fontenc} 
\hypersetup{
    colorlinks=false,
    linkcolor=black,
    citecolor=black,
    urlcolor=black,
    hidelinks
}

\author{Rajiv Teja Nagipogu}
\author{John H. Reif}
\email{rajivteja.nagipogu@duke.edu}
\affiliation{Department of Computer Science, Duke University, 2127 Campus Drive, Durham, NC 27708}
\title{Neural CRNs: A Natural Implementation of Learning in Chemical Reaction Networks}

\abbreviations{IR, NMR, UV}

\newcommand{\bs}[1]{\boldsymbol{#1}}
\newcommand{\mc}[1]{\mathcal{#1}}
\newcommand{\mb}[1]{\mathbb{#1}}
\newcommand{\mrm}[1]{$\mathrm{#1}$}

\newcommand{\la}[1]{\leftarrow}
\newcommand{\partfrac}[2]{\frac{\partial #1}{\partial #2}}

\newcommand{\fnlcls}[0]{f_\theta^{\mathrm{nlcls}}}

\newtheorem{definition}{Definition}%

\begin{document}

\maketitle

\abstract{Molecular circuits capable of autonomous learning could unlock novel applications in fields such as bioengineering and synthetic biology. To this end, existing chemical implementations of neural computing have mainly relied on emulating discrete-layered neural architectures using steady-state computations of mass action kinetics. In contrast, we propose an alternative dynamical systems-based approach in which neural computations are modeled as the time evolution of molecular concentrations. The analog nature of our framework naturally aligns with chemical kinetics-based computation, leading to more compact circuits. We present the advantages of our framework through three key demonstrations. First, we assemble an end-to-end supervised learning pipeline using only two sequential phases, the minimum required number for supervised learning. Then, we show (through appropriate simplifications) that both linear and nonlinear modeling circuits can be implemented solely using unimolecular and bimolecular reactions, avoiding the complexities of higher-order chemistries. Finally, we demonstrate that first-order gradient approximations can be natively incorporated into the framework, enabling nonlinear models to scale linearly rather than combinatorially with input dimensionality. All the circuit constructions are validated through training and inference simulations across various regression and classification tasks. Our work presents a viable pathway toward embedding learning behaviors in synthetic biochemical systems.}


\clearpage


\section{Introduction}\label{sec:introduction}

Learning is a key characteristic of all living organisms, enabling them to survive and function in constantly changing environments \cite{Koshland2002-jw}. In higher organisms, this adaptability is enabled through intricate neuronal networks that give rise to emergent properties such as memory, learning, and decision making. Remarkably, despite lacking such sophistication, single-celled organisms display a range of adaptive behaviors driven by their biochemical circuits. For example, bacteria navigate chemical gradients through chemotaxis \cite{webre2003bacterial}, slime molds exhibit intelligent foraging behaviors \cite{jabr2012brainless,nakagaki2000maze}, and Paramecia can develop an avoidance response through classical conditioning \cite{hennessey1979classical}. These examples suggest that intelligent behaviors can emerge solely from biomolecular interactions and regulatory dynamics, motivating efforts to functionally replicate such behaviors in engineered molecular systems \cite{hjelmfelt1991chemical}. If realized, these systems could enable novel applications in many areas of bioengineering, such as autonomous biosensing \cite{hua2022dna}, \textit{in situ} diagnostics \cite{zhang2020cancer}, and smart therapeutics \cite{zhang2021smart}. 

Recently, the field of \textit{molecular computing} has made significant strides in implementing programmed computation using synthetic biomolecular circuits. In particular, nucleic acid strand displacement motifs have been used to develop numerous computational devices, including Boolean logic gates \cite{qian2011simple, song2017renewable, de2007molecular, Xie2011-qh}, dynamical systems \cite{srinivas2017enzyme, chen2013programmable, montagne2011programming},  and even feedforward neural networks \cite{hjelmfelt1991chemical, Lakin2014-cb, Okumura2022-jm}. Similar to digital programming, these circuits are typically engineered by first specifying the desired behaviors using abstract chemical reaction networks (CRNs), which are then systematically compiled into concrete biomolecular implementations \cite{cook2009programmability}.

Despite this progress, designing adaptive molecular systems that can modify their behavior in response to environmental changes remains a significant challenge \cite{Lakin2014-cb,Lakin2016-al}. Embedding such adaptive capabilities would greatly enhance the versatility of molecular circuits, allowing them to be deployed in dynamical biochemical environments. Here, we propose the use of neural networks as a foundational paradigm to incorporate adaptability at the molecular scale. Neural networks are particularly well-suited for this role as: (i) they operate under mathematically grounded learning rules that can be translated into chemically feasible reaction schemes, and (ii) they provide compact parametric representations of functions, crucial for the low-resource paradigm of chemical computing. Although alternative paradigms of chemical learning exist (e.g., P-systems \cite{puaun2000computing}, reaction-diffusion systems \cite{turing1990chemical}, reservoir computing \cite{jaeger2001echo,maass2002real}), they often rely on specialized architectures, complex spatial organization, or external readout mechanisms, which are difficult to implement in well-mixed reactions.  

Prior implementations of chemical neural networks (CheNNs) have been implemented using a range of molecular mechanisms. Hjelmfelt \textit{et al.}\cite{hjelmfelt1991chemical} developed an enzymatic circuit mimicking a McCulloch-Pitts neuron \cite{mcculloch1943logical} and extended it into a binary perceptron system. Banda \textit{et al.}\cite{Banda2013Online,banda2014learning} introduced two chemical perceptron designs using analog multiplication as the core template for constructing simple online learning perceptron circuits. Similarly, Lakin \textit{et al.}\cite{Lakin2014-cb,Lakin2016-al} implemented two DNA-based perceptrons: the first using the strand-cleaving activity of DNAzymes \cite{gong2015dnazyme} and the second using a ``buffered'' DNA strand displacement motif \cite{zhang2011dynamic,qian2011scaling}. 
These perceptron designs were later  extended to multilayer networks \cite{Blount2017-ni} and enhanced with nonlinear activations in the hidden layer \cite{anderson2021reaction,Vasic2022-an, Arredondo2022-qj}. Parallel efforts have developed pattern recognition circuits using \textit{winner-take-all} computation \cite{Cherry2018-uz, xiong2022molecular} and nucleation-controlled DNA tile assembly \cite{evans2022pr}, and probabilistic models such as Markov chains \cite{singh2019reaction,Virinchi2018-lx} and Boltzmann machines \cite{poole2017chemical,poole2025autonomous} using stochastic CRNs.  

While these systems mark important progress, most still rely on chaining together chemical implementations of discrete algebraic primitives, requiring auxiliary modules to coordinate temporal staging. To address this issue, we propose \textit{Neural CRNs}, a general-purpose chemical neural network framework that models neural computations through the intrinsic concentration dynamics of chemical species. Our approach draws inspiration from \textit{Neural ordinary differential equations} (Neural ODEs) \cite{chen2018neural}, a dynamical system that models neural computations using the ODE dynamics of a set of state variables. Accordingly, the CRNs in the Neural CRNs framework are designed to emulate the dynamics of a reference Neural ODE system. This design allows chemical reactions within the framework to function as atomic end-to-end computational units, enabling scalable and practical implementations of chemical learning.  

The remainder of this manuscript is organized as follows. Section 2 introduces the Neural ODEs framework and the associated supervised learning procedure. Section 3 presents the construction of the Neural CRN architecture and its learning protocol, highlighting key modifications from the Neural ODE approach. Section 4 provides simulation results that demonstrate the ability of our framework to learn a variety of regression and classification tasks. Section 5 elaborates on key design decisions, provides an architectural comparison with prior work, and presents directions for future research. 

\section{Preliminaries}\label{sec:prelim}

\subsection{Neural Ordinary Differential Equations}

\begin{figure}[htbp]
    \centering
    \includegraphics[width=0.9\linewidth]{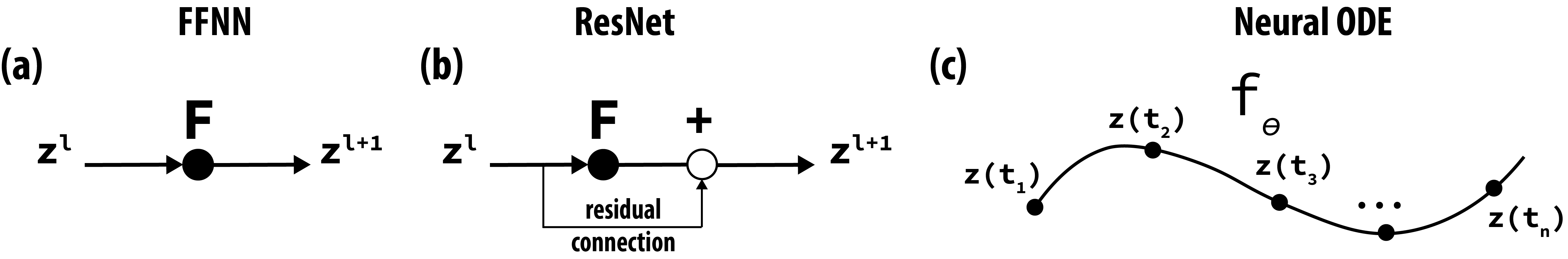}
    \caption{A comparison between the hidden state transformations in (a) vanilla Feedforward neural networks; (b) Residual Neural Networks (ResNets); and (c) Neural ODEs.}
    \label{fig:ffnn_node}
\end{figure}

Neural ODEs \cite{chen2018neural} are a class of dynamical system-based neural networks in which both the core neural computations--hidden state transformation during the forward pass and gradient computation during the backward pass--are modeled as continuous-time evolution of a set of state variables governed by ordinary differential equations (ODEs). Conceptually, Neural ODEs can be viewed as continuous-time analogs of residual neural networks (ResNets) \cite{Lu2018-vy}, a class of feedforward neural networks (FFNNs) that incorporate a `residual' connection between the input and output of each layer transformation (see Figure~\ref{fig:ffnn_node}b). As the number of layers approaches infinity, the discrete updates of a ResNet converge to a continuous trajectory of the network state (see Figure~\ref{fig:ffnn_node}c). Figure~\ref{fig:ffnn_node} illustrates this conceptual progression by comparing hidden state transformations across FFNNs, ResNets, and Neural ODEs. 

The architecture of a Neural ODE system can be specified using the following hyperparameters:

\begin{enumerate}
\item \textit{Dimensionality} $d$: The dimensionality of the hidden state vector $\bs z \in \mb R^d$; analogous to layer width in FFNNs.
\item \textit{Parameters} $\theta$: The learnable parameters of the Neural ODE system; analogous to weights and biases in FFNNs.
\item \textit{State dynamics function} $f_\theta$: Specifies the dynamics of $\bs z$; analogous to the layer transformations in FFNNs.
\item \textit{Time domain} $[t_i, t_f]$: The time interval over which $\bs z$ evolves; analogous to layer-depth in FFNNs.
\item \textit{Input projection matrix} $W_{\mathrm{in}}$: Initializes $\bs z$ by mapping the input $\bs x$ to $\bs z$ space $\bs z(t_i) = W_{\mathrm{in}} \bs x$; similar to the input layer in FFNNs.
\item \textit{Output projection matrix} $W_{\mathrm{out}}$: Projects the final hidden state to the output space: $\hat{y} = W_{\mathrm{out}} \bs z(t_f)$; similar to the output layer in FFNNs.
\end{enumerate}

%

\subsubsection*{Supervised learning in Neural ODEs}

Here, we briefly outline the supervised learning procedure in the Neural ODEs framework. For this, we consider an example learning task with input-output pairs $(\bs x, \bs y)$, where $\bs x \in \mb R^{d_x}$ and $\bs y \in \mb R^{d_y}$. For simplicity, we assume that the input $\bs x$ and the hidden state $\bs z$ have the same dimensionality ($d_x = d$), and that the output $\bs y$ is a scalar ($d_y = 1$).  Under these assumptions, the input projection reduces to an identity transformation ($W_{\mathrm{in}}$ = $\mc I_d$) and the output projection becomes a unit-weighted perceptron ($W_{\mathrm{out}} = \bs 1^\top$). Finally, we use $f_\theta = \theta \odot \bs z$ as the hidden state dynamics function.

The learning procedure in Neural ODEs follows the standard supervised learning protocol comprising the feedforward phase and the feedback (or learning) phase. In the \textit{feedforward} phase, the input is propagated through the system by evolving the hidden state $\bs z$ forward in time according to the dynamics defined by $f_\theta$. In the \textit{feedback} phase, parameter gradients $\bs g_\theta =  \nabla_\theta \mc L$ are calculated by evolving a gradient state backward in time using the \textit{adjoint sensitivity method} \cite{Pontryagin1987-qv}. 

\textbf{Feedforward phase.} The feedforward phase takes place over the time interval $[t_i, t_f]$. It begins by initializing the hidden state with the input $\bs x$ at $t=t_i$: $\bs z(t_i) = \bs x$ (recall that $W_{\mathrm{in}} = \mc I_d$).  The hidden state $\bs z$ then evolves according to $f_\theta$, where the final state $\bs z(t_f)$ is obtained by solving the resulting initial value problem (IVP) using standard numerical ODE solvers \cite{runge1895numerische,kutta1901beitrag}. Finally, since $W_{\mathrm{out}} = \bs 1^\top$, the predicted output $\hat{y}$ is calculated by adding the components of $\bs z(t_f)$: 
$$\hat{y} = \sum\limits_{i=1}^{d} z_i(t_f).$$ A summary of the feedforward phase is provided in Table~\ref{tab:node_fwd}. 

\begin{table}[htbp]
    \centering
    \footnotesize
    \caption{The feedforward phase evolves the hidden state dynamics over the time interval $[t_i, t_f]$, governed by the vector field $f_\theta$. At $t=t_i$, $\bs z$ is initialized to the input vector $\bs x$. The hidden state ODE is then evolved according to the user-picked $f_\theta$ function (in this case $f_\theta = \theta \odot \bs z$) until $t=t_f$. The final hidden state $\bs z(t_f)$ is evaluated by solving the corresponding IVP.} 
    \label{tab:node_fwd}
    \begin{tabular}{|c|c|c|c|}
        \hline
        \textbf{ODE} & \textbf{Initialization} (at $t=t_i$) & \textbf{For $f_\theta = \theta \odot \bs z$} & \textbf{IVP} \\
        \hline
        \rule{0pt}{4ex}$\dfrac{d\bs z}{dt} = f_\theta(\bs x, \bs z)$ & \rule{0pt}{4ex}$\bs z(t_i) = \bs x $ & \rule{0pt}{4ex}$\dfrac{d\bs{z}}{dt} = \theta \odot \bs{z}$ & \rule{0pt}{4ex}$\bs{z}(t_f) = \bs{z}(t_i) + \int\limits_{t_i}^{t_f} \theta \odot \bs z\ dt$ \\[2ex]
        \hline
    \end{tabular}
\end{table}

\textbf{Feedback phase.} The feedback phase is where the Neural ODEs fundamentally differ from conventional FFNNs. Instead of using the chain rule of backpropagation \cite{rumelhart1995backpropagation}, Neural ODEs employ a more efficient adjoint sensitivity method \cite{Pontryagin1987-qv} to compute parameter gradients. This approach introduces an auxiliary variable known as the \textit{adjoint state} $\bs a$, and uses it to specify the feedback phase dynamics of $\bs g_\theta$. These dynamics are interdependent: $\bs g_\theta$ evolves based on $\bs a$ and $\bs z$, and $\bs a$ evolves based on $\bs z$ ($\bs a(t) = \partfrac{\mc L}{\bs z}(t)$). Accordingly, the overall feedback phase is described by a coupled system of three ODEs---$\bs g_\theta$, $\bs a$, and $\bs z$ (see Table~\ref{tab:node_bwd}). These equations, collectively referred to as the \textit{feedback ODEs}, are simultaneously integrated backward in time from $t_f$ to $t_i$, to yield the final gradient state $\bs g_\theta(t_i)$.

Table~\ref{tab:node_bwd} summarizes the feedback phase. The first column shows the feedback ODEs in their most general form (see Chen \textit{et al.}\cite{chen2018neural} for their full derivation). The second, third, and fourth columns show how the corresponding state variables are initialized, evolved, and computed. The hidden state $\bs z$ takes on its final value from the end of the feedforward phase $\bs z(t_f)$ and is integrated backward in time to $\bs z(t_i)$. The adjoint state is initialized at $t=t_f$ as the partial derivative of the loss with respect to $\bs z$. Assuming a \textit{squared error} loss $\mc L_{sq} = \frac{1}{2}(\hat{y} - y)^2$ and $W_{out} = \bs 1^\top$, this expression evaluates to $\bs a(t_f) = \hat{y} - y$. The adjoint dynamics are then integrated backward in time to obtain $\bs a(t_i)$. Similarly, the gradient state $\bs g_\theta$ is initialized to $\bs 0$ at $t=t_f$ and evolved backward to obtain its final value $\bs g_\theta(t_i)$. These gradients are then used to update the parameters via gradient descent: $\theta^{\mathrm{new}} = \theta^{\mathrm{old}} - \eta \ \bs g_\theta(t_i)$, where $\eta$ is the learning rate parameter.

\begin{table}[htbp]
    \centering
    \footnotesize
    \caption{Feedback phase ODEs and their corresponding IVPs in the Neural ODEs supervised learning procedure. First column depicts the coupled ODE system corresponding to the backpropagation of $\bs z$, $\bs a$, and $\bs g_\theta$. Second column describes their initialization: (row 1) $\bs z$ is initialized to its final state at the end of the feedforward phase; (row 2)$\bs a$ is initialized by considering squared error ($\mc L_{sq} = \frac{1}{2}(\hat{y}-y)^2)$ as the loss function and calculating $\bs a(t_f) = \frac{\partial \mc L_{sq}}{\partial \bs z}(t_f)$. Since $\hat{y} = \sum_k z_k(t_f)$, the adjoint expression leads to $\bs a(t_f) = \hat{y} - y$; (row 3) The gradients are initialized to zero. Third column shows the feedback phase ODEs as per the chosen $f_\theta$. Finally, the fourth column depicts the ``backward-in-time'' IVPs of the three feedback ODEs.} 
    \label{tab:node_bwd}
    \begin{tabular}{|c|c|c|c|}
        \hline
        \textbf{ODE} & \textbf{Initialization (at $t=t_f$)} & \textbf{For $f_\theta = \theta \odot \bs z$} & \textbf{IVP} \\
        \hline
        \rule{0pt}{4ex}$\dfrac{d\bs{z}}{dt} = f_\theta(\bs x, \bs{z})$ & \rule{0pt}{4ex}$\bs{z}(t_f)$ & \rule{0pt}{4ex}$\dfrac{d\bs{z}}{dt} = \theta \odot \bs{z}$ & \rule{0pt}{4ex}$\bs{z}(t_i) = \bs{z}(t_f) + \int_{t_f}^{t_i} \theta \odot \bs z\ dt$ \\[2ex]
        \hline
        \rule{0pt}{4.5ex}$\dfrac{d\bs{a}}{dt} = -\bs{a}^\top \dfrac{\partial f_\theta}{\partial \bs{z}}$ & \rule{0pt}{4.5ex}$\bs{a}(t_f) = \hat{y} - y$ & \rule{0pt}{4.5ex}$\dfrac{d\bs{a}}{dt} = -\bs{a} \odot \theta$ & \rule{0pt}{4.5ex}$\bs{a}(t_i) = \bs{a}(t_f) + \int_{t_f}^{t_i} -\bs a\odot \theta \ dt$ \\[2ex]
        \hline
        \rule{0pt}{4.5ex}$\dfrac{d\bs{g}_\theta}{dt} = -\bs{a}^\top \dfrac{\partial f_\theta}{\partial \theta}$ & \rule{0pt}{4.5ex}$\bs{g}_\theta(t_f) = \bs 0$ & \rule{0pt}{4.5ex}$\dfrac{d\bs{g}_\theta}{dt} = -\bs{a} \odot \bs{z}$ & \rule{0pt}{4.5ex}$\bs{g}_\theta(t_i) = \bs{g}_\theta(t_f) + \int_{t_f}^{t_i} -\bs{a} \odot \bs z\ dt$ \\[2ex]
        \hline
    \end{tabular}
\end{table}

\subsection{From Neural ODEs to Neural CRNs}\label{ssec:kde}

The analog nature of Neural ODEs makes them a suitable reference system for designing the Neural CRNs framework. Specifically, a Neural CRN can be constructed by emulating the ODE dynamics of a reference Neural ODE system through mass action kinetics. However, not all Neural ODEs can be converted into Neural CRNs as mass action kinetics represents a restricted subclass of polynomial ODEs known as \textit{kinetic differential equations} (KDEs) \cite{hars1981inverse}. 

\begin{definition}
    \label{def:kdes}
    A KDE is a polynomial ODE that satisfies the following requirements.
    \begin{enumerate}
        \item All its variables must be positive-valued.
        \item The ODE must not have negative cross-effects, i.e., any term on the right-hand side with a negative sign must contain the differential variable.
    \end{enumerate}
\end{definition} We utilize the \textit{canonic mechanism}\cite{hars1981inverse} described in Algorithm S1 to translate a KDE into a CRN. In summary, this mechanism maps each term on the right-hand side of the KDE into a single chemical reaction: the factor variables of the term become the reactants, the differential variable becomes the product, and the coefficient becomes the rate constant. Moreover, the reaction will be catalytic in all of its reactants. For example, KDE: $\frac{dy}{dt} = kx_1 x_2$ translates into the reaction $X_1 + X_2 \xrightarrow[]{k} Y + X_1 + X_2$. 

However, not all ODEs in the Neural ODEs framework satisfy the KDE requirements. For example, the possibility of negative gradients or the backward integration of state variables during the feedback phase produces ODEs that violate them; we refer to such systems as \textit{non-KDEs}. In this work, we convert the non-KDEs into KDEs by reinterpreting their variables in the dual-rail form (see SI Text S4). Table~\ref{tab:ode_types} presents the CRN translations of common KDEs and non-KDEs used in our Neural CRN constructions. 

\begin{table}[h!]
    \centering
    \renewcommand{\arraystretch}{1.8} 
    \setlength{\tabcolsep}{10pt} 
    \begin{tabular}{|c|c|c|}
        \hline
        \textbf{Type} & \textbf{ODE} & \textbf{CRN} \\ \hline
        Type-I   & $\frac{dz}{dt} = xy$, \quad $x, y, z \in \mathbb{R}^+$ & $X + Y \to Z + X + Y$ \\ \hline
        Type-II  & $\frac{dz}{dt} = xy$, \quad $x, y, z \in \mathbb{R}$  & 4 Type-I reactions \\ \hline
        Type-III & $\frac{dz}{dt} = -xy$, \quad $x, y, z \in \mathbb{R}$ & 4 Type-I reactions \\ \hline
        Type-IV  & $\frac{dz}{dt} = z^q$, \quad $q\in \mathbb{Z}^+, z \in \mathbb{R}^+$ & $q Z \to (q+1) Z$ \\ \hline
        Type-V   & $\frac{dz}{dt} = -z^q$, \quad $q\in \mathbb{Z}^+, z \in \mathbb{R}^+$ & $q Z \to (q-1) Z$ \\ \hline
    \end{tabular}
    \caption{CRN translations of five polynomial ODE systems frequently used in the construction of Neural CRN circuits.}
    \label{tab:ode_types}
\end{table}

Notice that the ``backward time'' integrated feedback ODEs in the Neural ODEs framework cannot be translated directly into CRN dynamics because CRNs are physical processes and, therefore, cannot run backward in time. To resolve this issue, we invert the sign parities of the feedback ODEs by redefining their time variable $t$ as $\tau = t_f - t$. This change transforms the backward integration over $t \in [t_f, t_i]$ into a forward-time CRN evolution over $\tau \in [0, t_f-t_i]$. 

\section{Results}

In this section, we present the implementation of a supervised learning algorithm within the Neural CRNs framework. Before detailing the construction, we first introduce the assumptions and notational conventions used throughout the work.

\begin{enumerate}

    \item We use the lowercase alphabet ($x$) to represent scalar variables, the bold lowercase alphabet for vector variables ($\bs x$), and the uppercase alphabet ($X$) for chemical species.

    \item For notational simplicity, we occasionally use the variable $x$ as a shorthand for the concentration of the species $X$, when the context is clear. 

    \item Unless stated otherwise, chemical species are assumed to be specified in the \textit{dual-rail notation}, where a scalar variable $x$ is represented by the concentration difference of two complementary species: $x = [X^+] - [X^-]$. The $\pm$ superscripts on the species indicate positive or negative sign parity of the dual-rail species.

    \item The dual-rail species for a variable $x$ are initialized as follows.
    \begin{align*}
        [X^+] &= \max(0, x) \\
        [X^-] &= \max(0, -x).
    \end{align*}\label{assumption: dual_rail_init} We use the shorthand $X \xleftarrow[]{init} x$ to denote this initialization.

    \item We use phrases  such as \textit{the $Z$ species} or \textit{the hidden state species $Z$} to collectively refer to a species type. Similarly, the terms \textit{positive (negative)} $Z$ \textit{species} are used to represent the positive (negative) dual-rail species $Z$.

    \item The unspecified rate constants of chemical reactions are assumed to be unit-valued.

    \item We assume the presence of an oscillatory \textit{clock mechanism}, where the dominant signal during each oscillatory phase conditionally activates the corresponding set of reaction pathways. Since supervised learning is inherently sequential, this mechanism coordinates the transfer of control from one stage to the next.

    \item Although omitted for brevity, fast annihilation reactions between complementary dual-rail species are implicitly assumed throughout.

    \item Reactions labeled as `fast' are assumed to occur on a timescale much shorter than that of unlabeled `slow' reactions, and are therefore treated as instantaneous in comparison. 
\end{enumerate}

\subsection{Supervised learning in Neural CRNs}\label{ssec:ncrn_arch}

Here, we describe the supervised learning procedure in the Neural CRNs framework. As the running example, we consider a learning task involving two-dimensional inputs $\bs x \in \mb R^2$ and scalar outputs $y \in \mb R$, modeled after the Neural ODE system introduced earlier. This system uses $f_\theta = \theta \odot \bs{z}$ as the hidden state dynamics function, with the input and output projection matrices set to $W_{\mathrm{in}} = \mathcal{I}_2$ and $W_{\mathrm{out}} = \bs{1}^\top$, respectively. These choices fix the dimensionality of the key variables: $\bs{z}, \bs{a} \in \mathbb{R}^2$ and $\theta, \bs{g}_\theta \in \mathbb{R}^2$.

For simplicity, we assume that both inputs are positive and that the system should approximate a positive-valued function: $\bs x \in \mb R^2_{>0}$ and $y \in \mb R_{>0}$. This in turn allows $\theta$ and $\bs z$ to be positive throughout the process, allowing their corresponding species $P$ and $Z$ to be represented in the \textit{single-rail} notation.  Only adjoints $a_i$ and gradients $\bs g_\theta$, which can have negative values, are represented as dual-rail chemical species $A_i^\pm$ and $G_i^\pm$, respectively. Since we define a single supervised learning iteration, we set the run-time to $t_i = 0$ and $t_f = T$.

The supervised learning procedure in the Neural CRNs framework is divided into four computational stages, denoted \mrm{N1} through \mrm{N4}. Each stage corresponds to a key operation in the pipeline. Specifically, the \mrm{N1} stage evolves the hidden state species $Z$; the \mrm{N2} stage produces the adjoint species $A$ and the output species $\hat{Y}$; the \mrm{N3} stage accumulates the gradient species $G$; and finally, the \mrm{N4} stage updates the parameter species $P$ using the accumulated $G$ species. These distinct stages are allowed to coexist and are temporally coordinated within a single pot using an oscillatory clock mechanism. Here, the dominant clock signal during each oscillation is programmed to conditionally activate the reaction pathways corresponding to each stage, thereby enabling the time-multiplexed execution of the overall learning procedure.  Table~\ref{tab:ncrn_supervised} summarizes the role of each stage in the learning procedure and its associated CRN, while Figure~\ref{fig:neuralcrn_training} provides a visual overview of the overall training pipeline. We describe each stage in detail below. 

\begin{table}[htbp]
    \centering
    \renewcommand{\arraystretch}{1.8} 
    \setlength{\tabcolsep}{6pt} 
    \footnotesize
    \begin{tabular}{|c|c|c|c|c|c|}
        \hline
        \textbf{Stage} & \textbf{Clock Phase} & \textbf{Computation} & \textbf{CRN} & \textbf{Time Interval} & \textbf{Reactions} \\
        \hline
        \mrm{N1}  & C1 & $\dfrac{dz_i}{dt} = \theta_i z_i$ & \mrm{CRN_{f}^1} & $0 \to T$ &
        $P_{i} + Z_{i} \to Z_{i} + P_{i} + Z_{i}$ \\
        \hline
        \multirow{4}{*}{\mrm{N2}} & \multirow{4}{*}{C2} & $z_b = z$ & \multirow{4}{*}{\mrm{CRN_b^0}} & \multirow{4}{*}{$T \to T + \epsilon$} & $Z_i \to {Z_i^{b}} + A_1^+ + A_2^+ + \hat{Y}$ \\
        &  & $a = z_1 + z_2 - y$  &  &  & $Y \to A_1^- + A_2^-$ \\
        &  & $g_\theta = 0$  &  &  & $G \gets 0$ \\
        &  & $\hat{y} = z_1 + z_2$ (optional)  &  &  & \\
        \hline
        \mrm{N3} & C2 & See Table~\ref{tab:ncrn_bwd_all} & \mrm{CRN_b^1} & $T \to 2T$ & {See Table~\ref{tab:ncrn_bwd_all}} \\
        \hline
        \mrm{N4} & C1 & \begin{tabular}[c]{@{}c@{}} $\theta^{\mathrm{new}} = \theta^{\mathrm{old}} - \eta g_\theta$ \\ $z = x$ \end{tabular} & \mrm{CRN_b^2} & $2T \to 2T + \epsilon$ &
        \begin{tabular}[c]{@{}c@{}} $G^{\pm} \xrightarrow[]{k_1} P^{\mp}$ \\ $G^{\pm} \xrightarrow[]{k_2} \Phi$ \\ $\{Z, A, X\} \to \Phi$ \\ $X^{\mathrm{next}}\to Z$ \end{tabular} \\
        \hline
    \end{tabular}
    \caption{CRNs of the supervised learning procedure in Neural CRNs framework. The procedure consists of four stages (\mrm{N1}, \mrm{N2}, \mrm{N3}, and \mrm{N4}) organized into two clock phases (\mrm{C_1} and \mrm{C_2}). \mrm{N1} and \mrm{N3} are fast discrete stages, while \mrm{N2} and \mrm{N4} are slow analog stages. \mrm{N4} and \mrm{N1} run in clock phase $C_1$, while \mrm{N2} and \mrm{N3} run in clock phase $C_2$. \mrm{N1} stage models the evolution of the hidden state species ($Z$) and runs for a duration $T$. \mrm{N2} stage creates the adjoint species ($A$) and the backpropagating hidden state species ($Z^b$). This stage runs for a short duration $\epsilon\ (\ll T)$. \mrm{N3} stage creates the gradient species $G$ by evolving them alongside the feedback CRNs of $Z$ and $A$. This stage runs for a duration of $T$ and is discussed in more detail in Table~\ref{tab:ncrn_bwd_all}. \mrm{N4} stage updates the parameters, flushes out the non-parametric species ($Z^b, A, X$), and optionally feeds the next input ($X^{\mathrm{next}}$) into $Z$ species. The parameter update reactions involve a negative feedback of $G$ species into $P$ species ($\pm$ into $\mp$) alongside the decay of $G$ species, simulating subtraction. The values of $k_1$ and $k_2$ are determined by the learning rate $\eta$. This stage runs for a short duration $\epsilon$. }
    \label{tab:ncrn_supervised}
\end{table}

\noindent\textbf{Stage $\mathrm{\mathbf{N1}}$: Evolving the hidden state species (feedforward phase).} This stage models the feedforward dynamics of the hidden state species $Z$, emulating the forward phase ODE dynamics of the reference Neural ODE system. At time $t=0$, the input species $X_i$ are introduced into the system and copied into the corresponding $Z_i$ species. The $Z_i$ species then evolve according to \mrm{CRN_f^1}, which emulates the Type-I ODE $\frac{dz_i}{dt} = \theta_i z_i$ (see \mrm{N1} row in Table~\ref{tab:ncrn_supervised}). 

\noindent\textbf{Stage $\mathrm{\mathbf{N2}}$: Creating the adjoint species (transitioning into the feedback phase).}
This stage prepares the Neural CRN for the feedback phase by creating the adjoint species $A$. In a typical Neural ODE pipeline, the creation of adjoints involves three distinct steps: (i) output calculation $\hat{y} = \bs W_{\mathrm{out}} \odot \bs z$ , (ii) loss estimation $\mc L = \frac{1}{2} (\hat{y} - y)^2$, and (iii) adjoint computation $\bs a = \frac{\partial \mc L}{\partial \bs z}$. In our framework, we combine these three steps into a single arithmetic expression as follows:
\begin{gather}
    a_i = z_1 + z_2 - y.
\end{gather} The CRN for this expression is implemented by programming $Z_1$ and $Z_2$ to produce stoichiometrically equal amounts of $A^+$ species, and $Y$ to produce equal amounts of $A^-$ species (see row \mrm{N2} of Table~\ref{tab:ncrn_supervised}). In a feedforward-only Neural CRN system, this stage can be reconfigured to directly compute the output $\hat{y} = z_1 + z_2$, by programming $Z_1$ and $Z_2$ to produce the $\hat{Y}$ species. 

\noindent\textbf{Stage $\mathrm{\mathbf{N3}}$: Accumulating the gradient species (feedback phase).}
This stage represents the core computations involved in the feedback phase of supervised learning. Table~\ref{tab:ncrn_bwd_all} shows how the three feedback ODEs of $\bs z$, $\bs a$, and $\bs g_\theta$ in the Neural ODE framework are translated into the corresponding \textit{feedback CRNs} of $Z$, $A$, and $G$ species. The first row lists the feedback ODEs, originally defined to evolve backward in time. The second row inverts their signs, converting them into forward-time ODEs. The third row substitutes the $f_\theta$ expression into these ODEs. The fourth row decomposes them into individual differential terms. The fifth row categorizes these terms and specifies the appropriate translation template to convert them into CRNs (see Table~\ref{tab:ode_types}). Finally, the sixth row defines the initial concentrations of the state species $Z$, $A$, and $G$: (a) $Z$ species retain their concentrations from the end of the \mrm{N1} stage, (b) $A$ species are initialized at the end of the \mrm{N2} stage, and (c) $G$ species are initialized with zero concentration (no action required).
\begin{table}[htbp]
    \centering
    \footnotesize
    \renewcommand{\arraystretch}{2.5} 
    \setlength{\tabcolsep}{6pt} 
    \begin{tabular}{|c|c|c|c|c|}
        \hline
        \multirow{2}{*}{\textbf{Row}} & \multirow{2}{*}{\textbf{Remark}} & \multicolumn{3}{c|}{\textbf{Backpropagation}} \\ \cline{3-5}
        &  & \textbf{Hidden State} & \textbf{Adjoint} & \textbf{Gradient} \\
        \hline
        1 & $T \to 0$ & $\frac{d\bs z}{dt} = \theta \odot \bs z$ & $\frac{d\bs a}{dt} = -\bs a^\top \frac{\partial f_\theta}{\partial \bs z}$ & $\frac{d\bs g_\theta}{dt} = -\bs a^\top \frac{\partial f_\theta}{\partial \theta}$ \\
        \hline
        2 & $0 \to T$ & $\frac{d\bs z}{dt} = -\theta \odot \bs z$ & $\frac{d\bs a}{dt} = \bs a^\top \frac{\partial f_\theta}{\partial \bs z}$ & $\frac{d \bs g_\theta}{dt} = \bs a^\top \frac{\partial f_\theta}{\partial \theta}$ \\
        \hline
        3 & Expression & $\frac{d\bs z}{dt} = -\theta \odot \bs z$ & $\frac{d\bs a}{dt} = \bs a \odot \theta$ & $\frac{d\bs g_\theta}{dt} = \bs a \odot \bs z$ \\
        \hline
        4 & Expansion & $\frac{dz_i}{dt} = -\theta_{i}z_i$ & $\frac{da_i}{dt} = a_i\theta_{i}$ & $\frac{dg_{i}}{dt} = a_iz_i$ \\
        \hline
        5 & Translation & Type-III ODE & Type-II ODE & Type-II ODE \\
        \hline
        6 & Initialization & $Z \xleftarrow[]{init} \bs z(T)$ & $A \xleftarrow[]{init} z_1 + z_2 - y$ & $G \xleftarrow[]{init} 0$ \\
        \hline
    \end{tabular}
    \caption{\footnotesize Feedback CRNs of the Neural CRNs framework corresponding to $Z$ and $A$, and $G$ species. \textit{(Row 1)} Feedback ODEs of the Neural ODEs framework that run in reverse-time (from $t=T$ to $t=0$). \textit{(Row 2)} ODEs with their parities inverted to facilitate conversion into CRNs. \textit{(Row 3)} ODEs obtained by substituting the expression for $f_\theta$. \textit{(Row 4)} Terms in the ODE expansion. \textit{(Row 5)} Template from Table~\ref{tab:ode_types} to be applied for the ODE-to-CRN translation. \textit{(Row 6)} Initialization of the evolving species.}
    \label{tab:ncrn_bwd_all}
\end{table}

\noindent\textbf{Stage $\mathrm{\mathbf{N4}}$: Updating the parameter species (transitioning into the feedforward phase).}
This stage marks the end of a learning iteration, where the $G$ species accumulated during the \mrm{N3} stage are used to update the parameter species $P$. This update follows the gradient descent formulation ($\theta^{\mathrm{new}} = \theta^{\mathrm{old}} - \eta \bs g_\theta$) comprising two algebraic operations: (i) rational multiplication of $\eta$ and $\bs g_\theta$ to obtain $\Delta \theta = \eta \bs g_\theta$  and (ii) subtraction of $\Delta \theta$ from current parameters $\theta$. Here, for simplicity, we set $\eta = 1$ so that this combined computation can be implemented using a subtraction CRN (see SI Text S4), where the $G^-$ species feed positively and the $G^+$ species feed negatively into the $P$ species through the reactions $G^- \xrightarrow[]{} P$ and $G^+ + P \xrightarrow[]{} \Phi$. We discuss alternative CRN formulations for the case where $\eta < 1$ in SI Text S4. In addition to the parameter update, this stage also flushes the non-parametric and non-gradient species $Z$, $A$, and $X$ using the decay reactions $\{Z, A, X\}\rightarrow \Phi$, to reset the system for the next learning iteration.

\subsubsection*{Two-phase supervised learning pipeline}

The four-stage supervised learning procedure outlined above typically requires a four-phase clock mechanism, wherein the dominant clock signal in each phase selectively activates the CRN of the corresponding stage \cite{vasic2020crn++}. A higher number of sequential stages not only necessitates more clock phases but also increases the risk of unintended crosstalk between different stages, substantially raising the circuit complexity. To mitigate these issues, we present a streamlined supervised learning pipeline in which the aforementioned four stages are programmed to execute within a two-phase clock mechanism. 

This optimization is enabled by strategically leveraging time-scale separation. Observe from Table~\ref{tab:ncrn_supervised} that the four stages alternate between exclusively analog and discrete algebraic computations. The stages \mrm{N1} and \mrm{N3} are `analog', while \mrm{N2} and \mrm{N4} are `discrete'. Furthermore, all the discrete computations are modeled using unimolecular reactions, whereas analog computations are implemented using bimolecular or higher-order reactions. By enforcing time-scale separation between the unimolecular and higher-order reactions, the discrete stages could be executed quickly in a short time window $\epsilon\ (\ll T)$. They could then be triggered concurrently with their subsequent analog stages, which operate over a duration $T$. Specifically, \mrm{N2} overlaps with \mrm{N3} and \mrm{N4} overlaps with \mrm{N1} of the subsequent training iteration. Finally, by synchronizing the duration of the analog stages $T$ with the duration of the clock cycle, the entire supervised learning procedure could be executed within two clock phases. 

\begin{figure}[htbp]
    \centering
    \includegraphics[width=0.5\textwidth]{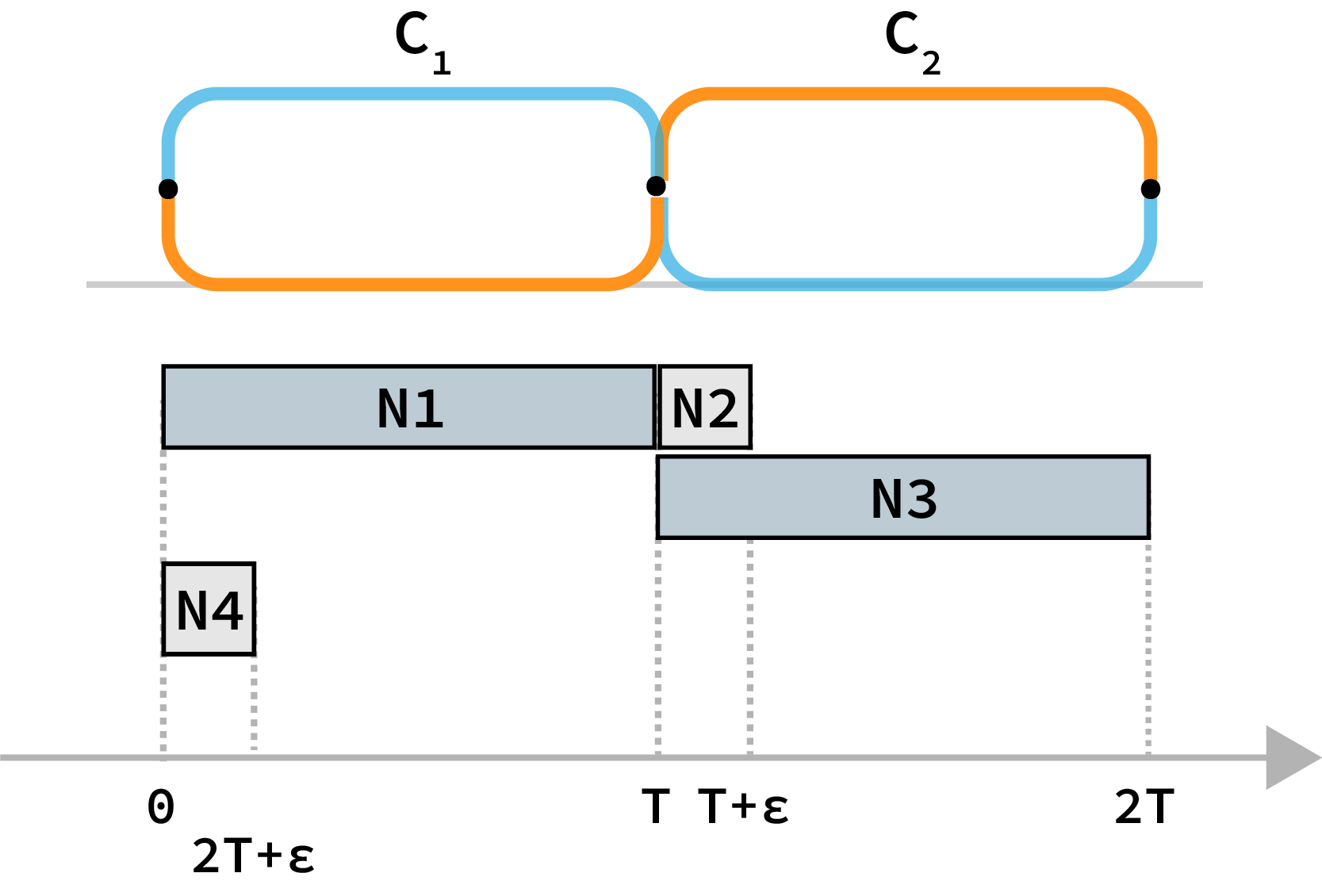}
    \caption{Arrangement of different stages in the Neural CRN supervised learning procedure on a timeline, illustrating their relative durations and the clock phases they run in.}
    \label{fig:stages_arrangement}
\end{figure}

Figure~\ref{fig:stages_arrangement} illustrates this alignment. Assume a two-phase clock mechanism with the clock phases labeled $C_1$ and $C_2$. The clock cycle begins with the $C_1$ phase at $t=0$, activating the \mrm{N1} stage. At $t=T$ the clock switches to the $C_2$ phase, deactivating the \mrm{N1} stage, and activating both the \mrm{N2} and \mrm{N3} stages. Since \mrm{N2} is a discrete stage, it finishes quickly within a short duration $\epsilon$ (from $t=T$ to $t=T+\epsilon$), while \mrm{N3} runs for the full duration $T$ (from $t=T$ to $t=2T$). At $t=2T$, the clock reenters the \mrm{C_1} phase, activating the \mrm{N4} stage and the \mrm{N1} stage of the next learning iteration (corresponding to a different input-output pair). \mrm{N4} runs for a short duration $\epsilon$ (from $t=2T$ to $t=2T+\epsilon$) to update the parameter species, while \mrm{N1} continues to process the next input for the full duration $T$. The clock phases associated with each stage are summarized in Table~\ref{tab:ncrn_supervised}.

\begin{figure}[htbp]
    \centering
    \caption{A schematic of the Neural CRN training procedure depicting the flow of information. Each training iteration consists of four stages \mrm{N1}, \mrm{N2}, \mrm{N3}, and \mrm{N4} separated by two clock triggers $C_1$ and $C_2$. Time-scale separation is enforced to separate them into slow `analog' and fast `discrete' stages. The `dashed' line in the right edge of a panel signifies that the corresponding stage overlaps runs in conjunction (albeit for a short duration $\epsilon$) with its adjacent stage.  At time $t=0$, \mrm{N1} begins with its $Z$ species initialized with input species $X$, which then evolve until $t=T$ according to \mrm{CRN_f^1}. At $t=T$, the $C_2$ phase is triggered, which starts both \mrm{N2} and \mrm{N3} stages. \mrm{N2} creates the adjoint species $A$ and hidden state backpropagation species $Z^b$ by running \mrm{CRN_b^0}. This stage lasts for a short duration till $t=T+\epsilon$. \mrm{N3} stage simultaneously executes the three backpropagation CRNs of $Z^b$, $A$, and $G$ species and runs for a duration of $T$ until $t=2T$. At $t=2T$, \mrm{C_1} triggers, switching the execution to both \mrm{N1} and \mrm{N4} stages. \mrm{N4} stage uses the $G$ species from the end of \mrm{N3} stage to update the parameter species $P$ by running \mrm{CRN_b^2} for a short duration $\epsilon$ till $t=2T + \epsilon$. Note that this phase overlaps with the \mrm{N1} stage of the next iteration.}
    \includegraphics[width=\textwidth]{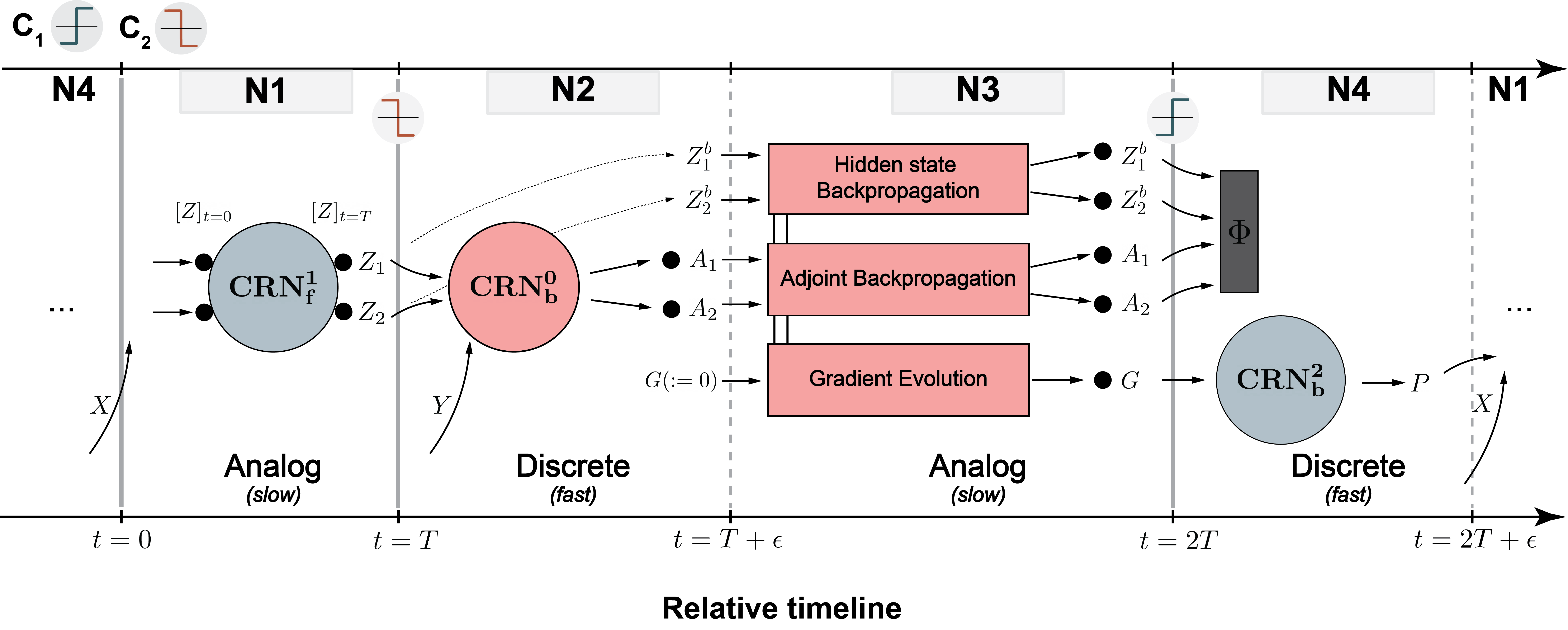}
    \label{fig:neuralcrn_training}
\end{figure}

\noindent\textbf{Avoiding crosstalk between simultaneously running stages.} An important consideration when concurrently executing sequential computations is to ensure that there are no \textit{cross-effects} between stages---specifically, ensuring that species produced in one stage are not consumed in another \cite{Cardelli2020-gf}. Notice in Table~\ref{tab:ncrn_supervised} that there is one such cross-effect involving the $Z$ species between the \mrm{N1} and \mrm{N4} stages. Specifically, $Z$ is a product in the \mrm{N1} stage and also a reactant in the \mrm{N4} stage (due to decay reactions). To neutralize this cross-effect, we preemptively copy the $Z$ species from the end of the \mrm{N1} stage into ``proxy'' feedback species $Z^b$ during the \mrm{N2} stage. The circuit then utilizes $Z^b$ in place of $Z$ during the \mrm{N3} and \mrm{N4} stages. Figure~\ref{fig:neuralcrn_training} illustrates this optimized two-phase supervised learning pipeline.

\subsection{Demonstrations of Supervised Learning in Neural CRNs} \label{ssec:lin_reg}

In the following sections, we validate our Neural CRNs framework by demonstrating that it can learn a range of regression and classification tasks. In \textit{regression} tasks, the model optimizes the network parameters to minimize the prediction error, while in \textit{classification} tasks, it optimizes the class discriminant function to maximize the accuracy of class assignment.

\subsection{Neural CRNs for Regression}

\subsubsection*{Neural CRNs for linear regression}
\textit{Linear regression} refers to the class of regression tasks in which the output is computed as a linear function on the input. Here, we present a Neural CRN circuit for linear regression termed the \textit{Linear Regressor Neural CRN} (LR-NCRN). We employ $f_\theta^{\mathrm{linreg}}$ as the hidden state dynamics function:
\begin{gather}
    f_\theta^{\mathrm{linreg}} = \theta \odot \bs x + \beta \label{eq:ftheta_linreg}
\end{gather} where $\theta \in \mb R^2_{>0}$ are the learnable parameters and $\beta \in \mb R_{>0}$ is the constant bias term. A supplementary objective of this construction is to showcase a minimal chemical learning circuit in the Neural CRNs framework.

\textbf{Dataset.} We created a synthetic linear regression dataset \textit{LinReg2D} to validate the LR-NCRN. This dataset consists of two-dimensional positive inputs $\bs x \in \mb R^2_{>0}$ and positive scalar outputs $y \in \mb R_{>0}$. The inputs are sampled from a uniform distribution $\bs x \sim \mc U(1.0, 5.0)$ and the corresponding outputs are generated according to the linear function:
\begin{gather}
    y = k_1 x_1 + k_2x_2 + k_0 + \xi
\end{gather} where the coefficients $k_1 = 1$, $k_2 = 2$, and the bias term $k_0=1$ are arbitrarily chosen, and $\xi \sim \mc N (0.0, 0.4)$ denotes the additive Gaussian noise. 

Since both inputs and outputs are strictly positive, it is possible for both $\theta$ and $\bs z$ to remain positive throughout the training, provided that the gradient step sizes for $\theta$ remain sufficiently low. As a result, the $P, Z, X, Y,$ and $\hat{Y}$ species are specified using single-rail representations, while the $A$ and $G$ species are represented using dual-rail encoding.

We detail the LR-NCRN circuit in Table~\ref{tab:lin_reg}---the left subtable lists the circuit species and the right subtable lists the circuit reactions. The \mrm{N1} stage involves three reactions: two from the Type-I ODE terms $\theta_ix_i$ and one from the $\beta$ term. The \mrm{N2} stage also involves three reactions, with the two $Z_i$ species feeding positively and the $Y$ species feeding negatively into the $A_i$ species. The \mrm{N3} stage involves only the feedback CRN of $G$ species and doesn't require running the feedback CRNs of $A$ and $Z$ species due to the lack of a $\bs z$-term in $f_\theta$ ($\frac{d\bs a}{dt} = \bs 0$; no $\bs z$-term in either $\frac{d\bs g}{dt}$ or $\frac{d\bs a}{dt}$). Accordingly, this stage requires four reactions, two per each Type-II ODE term $\frac{dg_i}{dt} = a_i x_i$. The \mrm{N4} stage, assuming $\eta = 1$, involves four reactions with the two $G_i^-$ species adding to the corresponding $P_i$ species concentrations and the two $G_i^+$ species reducing the $P_i$ species concentrations stoichiometrically. Overall, the LR-NCRN circuit utilizes 17 species and 14 reactions. The decay reactions of $A$ and $X$ from the end of each iteration (in \mrm{N4}) are excluded from the count to facilitate a fair comparison with similar prior circuits. Additional implementation details for this circuit are provided in SI Text S5.

\begin{table}[htbp]
    \centering
    \footnotesize
    \renewcommand{\arraystretch}{1.4} 
    \setlength{\tabcolsep}{10pt} 

    \begin{tabular}{c c c}
        \begin{tabular}[t]{|c|c|}
            \hline
            \textbf{Group name} & \textbf{Species} \\
            \hline
            Inputs & $X_1, X_2$ \\ \hline
            Output & $Y$ \\  \hline
            Parameters & $P_1, P_2$ \\ \hline
            States & $Z_1, Z_2$ \\ \hline
            Adjoints & $A_1^\pm, A_2^\pm$ \\ \hline
            Gradients & $G_1^\pm, G_2^\pm$ \\ \hline
            Clock signals & $C_1$, $C_2$ \\  \hline
            \textbf{Total} & 17 \\ \hline
        \end{tabular}
        & \quad &
        \begin{tabular}[t]{|c|c|c|}
            \hline
            \textbf{Stage} & \textbf{ODE} & \textbf{Reactions} \\
            \hline
            $\mathrm{N1}$ & $\frac{dz_i}{dt} = \theta_i x_i + \beta$ &
            \begin{tabular}{@{}c@{}}
                $X_i + P_i \to Z_i + X_i + P_i$  \\
                $B \to Z_1 + Z_2$
            \end{tabular} \\
            \hline
            $\mathrm{N2}$ & $a_i = z_1 + z_2 - y$ &
            \begin{tabular}{@{}c@{}}
                $Z_i \to A_{1}^+ + A_{2}^+$  \\
                $Y \to A_{1}^- + A_{2}^-$
            \end{tabular} \\
            \hline
            $\mathrm{N3}$ & $\frac{dg_i}{dt} = a_i x_i$ &
            $X_i + A_{i}^\pm \to G_{i}^\pm + X_i + A_i^\pm$ \\
            \hline
            $\mathrm{N4}$ & $\theta_i = \theta_i -g_i$ &
            \begin{tabular}{@{}c@{}}
                $G_i^- \to P_i$  \\
                $G_i^+ + P_i \to \Phi$
            \end{tabular} \\
            \hline
            \multicolumn{2}{|c|}{\textbf{Total}} & 14 \\ \hline
            \multicolumn{2}{|c|}{Flushing at iteration end} &
            $\{A_i^\pm, X_i\} \to \phi$ \\
            \hline
        \end{tabular}
    \end{tabular}
    \caption{The Linear Regressor Neural CRN (LR-NCRN) circuit designed for the LinReg2D task ($i = \{1, 2\}$). (\textit{left}) Species of the circuit. Since the task features positive-valued inputs and outputs, only $A$ and $G$ are required to be in dual-rail notation. Rest of the species are specified in single-rail notation. The circuit requires 17 species in total. (\textit{right}) Reactions in the resulting circuit. \mrm{N1} stage involves three reactions with two Type-I ODEs and one bias term. \mrm{N2} stage involves three reactions with two $Z$ species positively feeding and one $Y$ species negatively feeding into the $A$ species. \mrm{N3} stage only involves only gradient evolution (the other two `backpropagation' CRNs can be ignored) resulting in four reactions. \mrm{N4} stage involves negative feedback from $G$ species resulting in four reactions. Optionally at the end, the non-parametric species should be flushed out, preparing for the next iteration. The resulting circuit involves 14 reactions in total (excluding the flushing reactions at the end).}
    \label{tab:lin_reg}
\end{table}

\begin{figure}[htbp]
    \centering
    \includegraphics[width=\textwidth]{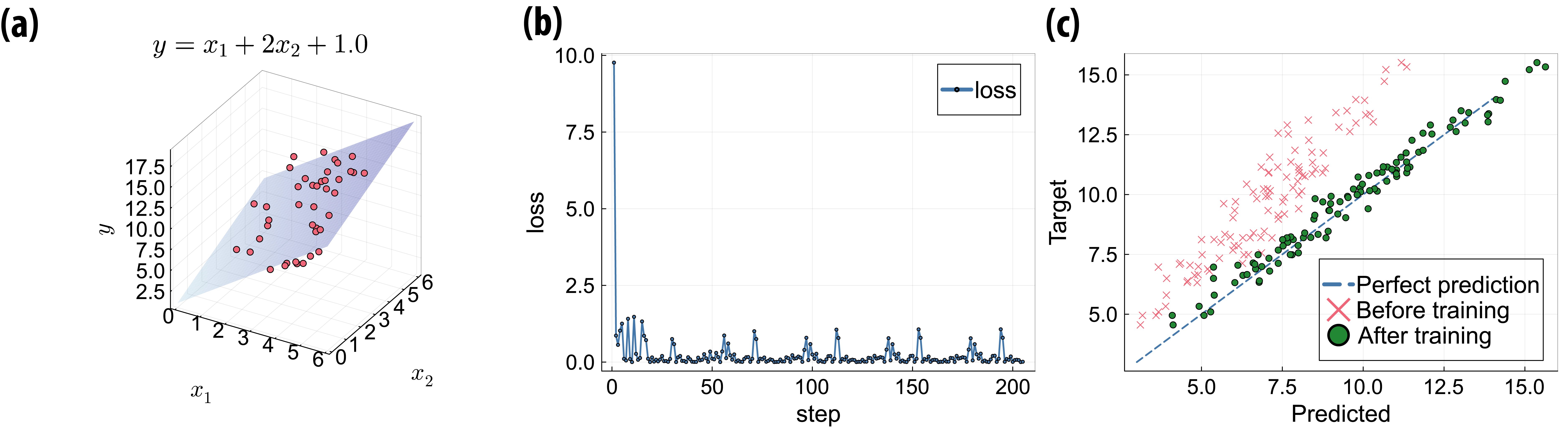}
    \caption{Results of training and inference of the Linear Regressor Neural CRN (LR-NCRN) on a linear regression task. (a) The training dataset juxtaposed with the output surface: $y = k_1x_1 + k_2x_2 + k_0$, with $k_1 = 1.0$, $k_2 = 2.0$, and $k_0 = 1.0$. (b) Step-wise loss at each iteration during training depicting loss convergence. (c) A comparison of predicted outputs on the test set against the target outputs before and after training.  The data points after training are closer to the ``perfect prediction line'' (on which \mrm{predicted = target}) demonstrating the ability of our Linear Regressor Neural CRN to model linear regression tasks.}
    \label{fig:lin_reg}
\end{figure}

Figure~\ref{fig:lin_reg} presents the learning behavior and predictive performance of the LR-NCRN circuit on the LinReg2D dataset. Figure~\ref{fig:lin_reg}a shows a scatter plot of the training set $(x_1, x_2, y)$ juxtaposed with the fitted surface plane $g(x_1, x_2) = k_1 x_1 + k_2x_2 + k_0$ ($k_1 = 1.0$, $k_2 = 2.0$, and $k_0 = 1.0$). Figure~\ref{fig:lin_reg}b shows the value of the loss function at each step, which demonstrates convergence during training. Figure~\ref{fig:lin_reg}c compares the $(\mathrm{predicted, target})$ points before and after training against the ``perfect prediction line'' ($\mathrm{prediction = target}$). The predictions correlate more closely with the targets after training than before, confirming that our LR-NCRN model can learn linear regression tasks.

\textbf{A minimal learning circuit in the Neural CRNs framework. } The LR-NCRN circuit described above can be further optimized to create a minimal learning circuit in the Neural CRNs framework. For example, considering a regression task with $k_0 = 0$, the bias term $\beta$ in $f_\theta$ could be safely removed, which allows removal of $B \xrightarrow[]{} Z_1 + Z_2$.  Furthermore, since the adjoint variables $a_i$ are constant and identical throughout, the four $A_i^\pm$ species can now be represented using only two species $A^+$ and $A^-$.  These optimizations eliminate two species and one reaction, yielding a circuit with 15 species and 13 reactions.

\subsubsection*{Neural CRNs for nonlinear regression}\label{ssec:nonlin_reg}

\textit{Nonlinear regression} refers to the class of regression tasks in which the output is modeled as a nonlinear transformation of the input. Here, we present a Neural CRN circuit to model nonlinear regression tasks, termed the \textit{Nonlinear Regressor Neural CRN} (NLR-NCRN). We use this construction to additionally illustrate the architectural modifications required to adapt the Neural CRNs framework for nonlinear modeling.


Nonlinear modeling in ODE-based neural networks like Neural ODEs or Neural CRNs differs significantly from the conventional FFNN way of applying nonlinear activation functions in the hidden layers. In particular, simply using a nonlinear $f_\theta$ is not sufficient to model nonlinearities. This is because the computations in these systems are represented as spatial flow trajectories in input space, uniquely defined by their initial input, meaning trajectories cannot intersect with themselves or others \cite{birkhoff1927dynamical}. Training them involves finding the right set of parameters that yield nonlinear and non-intersecting trajectories that correctly transform the inputs to outputs. Often, this is challenging as the desired transformation may lead to entangled spatial trajectories, leading to a strenuous training process (e.g., training a classifier on the Rings2D dataset described below). 

To counteract this issue, Neural ODEs incorporate a well-known machine learning technique known as \textit{implicit lifting} \cite{mercer1909xvi}, where a nonlinear task in the lower-dimensional space is converted into a linear task in the higher-dimensional space by applying a nonlinear kernel transformation on the input. For example, in the case of the Rings2D dataset, the two rings can be pushed by different amounts along the $z$ direction, converting the nonlinear classification task in 2D into a linear classification task in 3D. This process is typically referred to as \textit{augmenting} \cite{dupont2019augmented}. 

Integrating implicit lifting into the Neural ODE architecture involves three key modifications \cite{dupont2019augmented}: (i) padding the inputs with $p$ extra dimensions, (ii)  adjusting the dimensionality of the state variables accordingly, and (iii) employing a nonlinear $f_\theta$ in the ``augmented'' space. Notice that these modifications have the cumulative effect of applying a nonlinear kernel transformation on the input into a higher-dimensional space. Analogously, we adopt these modifications into the Neural CRN architecture by: (a) introducing $p$ additional input species, (b) appropriately expanding the state species $P, Z, A, G$, and (c) modifying the CRNs by using a nonlinear $f_\theta$ function in the augmented space ($\mb R^{d+p}$):

\begin{gather}
    f_{\theta}^{\mathrm{nlreg}} = \theta \odot \bs x - \bs z \odot \bs z \label{eq:ftheta_nlreg}
\end{gather} where $\bs x \in \mb R^{d+p}$, $\theta \in \mb R^{d+p}$, and $\bs z\in \mb R^{d+p}$.

\textbf{Dataset.} We created a synthetic nonlinear regression dataset \textit{NonLinReg2D} to validate the NLR-NCRN circuit. This dataset contains two-dimensional positive inputs $\bs x \in \mb R^2_{>0}$ and scalar positive outputs $y \in \mb R_{>0}$. The inputs are sampled from a uniform distribution $\bs x \sim \mc U \mathrm{(0.5, 2.0)}$ and the corresponding outputs are generated according to:
\begin{gather}
    y = x_1x_2 + x_2^2  + \xi
\end{gather} where $\xi \sim \mc N(0, 0.1)$ represents the Gaussian noise.

\begin{figure}[htbp]
    \centering
    \includegraphics[width=\textwidth]{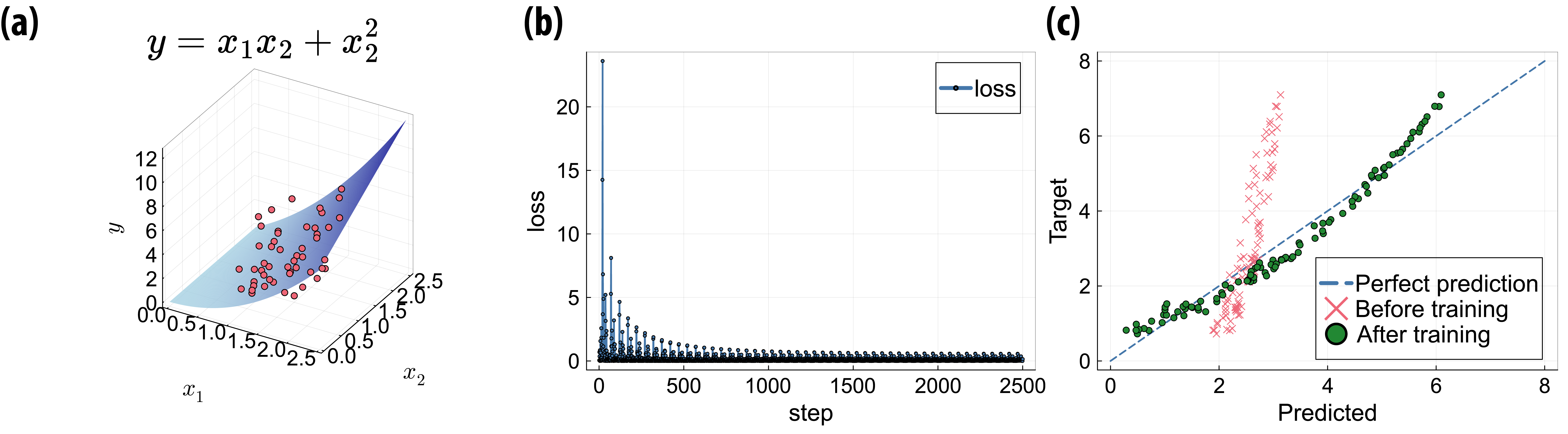}
    \caption{Results of training and inference of the NLR-NCRN model on a nonlinear regression task. (a) The training dataset juxtaposed against the output surface: $y = x_1x_2 + x_2^2$. (b) Step-wise loss at each iteration during training depicting loss convergence. (c) Comparison of predicted outputs on the test set against the target outputs before and after training.  The data points after training are closer to the ``perfect prediction line'' (on which \mrm{predicted = target}) demonstrating the ability of our NLR-NCRN circuit to model linear regression tasks.}
    \label{fig:nonlin_reg}
\end{figure}

Figure~\ref{fig:nonlin_reg} presents the results of training and inference by the NLR-NCRN on this dataset. Figure~\ref{fig:nonlin_reg}a visualizes the training data $(x_1, x_2, y)$ in a 3D space along with the fitted surface $y = x_1x_2 + x_2^2$. Figure~\ref{fig:nonlin_reg}b illustrates the loss convergence during training. Figure~\ref{fig:lin_reg}c contrasts the $(\mathrm{predicted, target})$ points before and after training against the perfect prediction line ($\mathrm{prediction = target}$). The predictions correlate more closely with the targets after training, demonstrating that our NLR-NCRN model can learn nonlinear regression tasks. To further establish its robustness, we evaluate the NLR-NCRN circuit on another nonlinear regression task $y = \sin (x_1) + x_2^2$ in SI Text S6.

\textbf{NLR-NCRN using first-order gradient approximation. }The $\bs z$-term in $f_\theta^{\mathrm{nlreg}}$ leads to autocatalytic behavior in the NLR-NCRN circuit during the feedback phase, leading to training instabilities. Although this autocatalytic growth can be partially mitigated---for example, by applying a damping factor $\alpha < 1$ to the $\bs z$-term or by reducing the evolution time $T$---the circuit remains sensitive to background noise, which can amplify exponentially if left unchecked. For robustness, the circuit should be designed with as few autocatalytic reactions as possible. We propose the following simplification to this end. Observe that the autocatalytic reactions in the circuit are mainly due to the feedback CRNs of $A$ and $Z$. What if we assume that the $A$ species remain unchanged throughout the feedback phase? First, it eliminates the feedback CRN of $A$. Furthermore, since gradient dynamics ($\frac{dg_i}{dt} = a_i x_i$) does not involve a $z$-term, the feedback CRN of $Z$ can also be eliminated. As a result, $G$ species evaluate to a final concentration value $g_i =Ta_ix_i$ (both $a_i$ and $x_i$ are constant). Since $a_i = \hat{y}-y$ (see Table~\ref{tab:ncrn_supervised}), this expression closely resembles the perceptron weight update rule\cite{rosenblatt1958perceptron} $\Delta w = \alpha (\hat{y} - y)x_i$ (scaled by a factor of $T$), implying that the simplification of maintaining constant adjoints is equivalent to computing first-order gradient approximations. We validate this simplified NLC-NCRN circuit in SI Text S6.

\subsection{Neural CRNs for Binary Classification}\label{ssec:ncrn_classification}
The regression architecture described above can be trivially repurposed into a binary classification architecture, by interpreting the network output $\hat{y}$ as a discriminant for classification. Specifically, the class labels for each input $\bs x$ are assigned by comparing $\hat{y}$ with a predefined threshold $\phi$ as shown below:
\begin{gather}
    \mathrm{label} = \begin{cases}
        \mathrm{ON} & \hat{y} > \phi \\
        \mathrm{OFF} & \text{otherwise}
    \end{cases}
\end{gather} During training, the network aims to minimize the prediction error between $\hat{y}$ and a real-valued target $y^{\mathrm{ON}}$ or $y^{\mathrm{OFF}}$, custom defined for each classification task. Below, we describe the construction of binary classifier Neural CRN circuits for both linearly separable and nonlinearly separable classes.

\subsubsection*{Linearly separable classes}

We construct the \textit{Linear Classifier Neural CRN} (LC-NCRN) circuit for modeling binary classification tasks with linearly separable classes. This circuit is assembled using $f_\theta^{\mathrm{lincls}}$ as the dynamics function.
\begin{gather}
    f_{\theta, \beta}^{\mathrm{lincls}}(\bs x, \bs z) = \theta \odot \bs x + \beta. \label{eq:ftheta_lincls}
\end{gather} At the end of the feedforward phase, the circuit output coincides with that of a perceptron output scaled by a factor $T$: $\hat{y}^{\mathrm{lincls}} = T(\theta \odot \bs x + \beta)$, confirming its linear nature. We validated this circuit using the Linear2D dataset.

\textbf{Dataset.} \textit{Linear2D} is a synthetic binary classification dataset with two-dimensional inputs $\bs x\in \mb R^2$ divided into equally populated linearly separable classes (see Figure~\ref{fig:cls_results}a). The inputs are sampled from a uniform distribution $\bs x \sim \mathrm{Uniform(0, 2)}$ and the class labels are assigned using the discriminant function $g^{\mathrm{lincls}}(x_1, x_2)= k_1x_1 + k_2x_2$:
\begin{gather*}
    y = \begin{cases}
        y^{\mathrm{ON}} & g^{\mathrm{lincls}}(x_1, x_2) > \phi \\
        y^{\mathrm{OFF}} & \text{otherwise}
    \end{cases}
\end{gather*} The classification parameters are set as follows: $k_1 = 1.0$, $k_2 = 2.0$, $y^{\mathrm{ON}} = 4$, $y^{OFF} = 0$, and $\phi = 2$.

Figure~\ref{fig:cls_results}a shows the training set, illustrating the distribution of data between the two classes. Figure~\ref{fig:cls_results}b presents the loss convergence curves in the training and validation sets, demonstrating that the LC-NCRN circuit can learn linear classification tasks. Figure~\ref{fig:cls_results}c shows the trained model's predictions on a uniformly spaced test grid, highlighting the learned decision boundary and the misclassified points (marked with an $\times$). The observed decision boundary successfully approximates the intended discriminant function $x_1 + 2x_2 = 2$, as evidenced by its intercepts on: (a) \textit{$x_1$-axis}: at $x_2 = 0.0$, $x_1 \approx 2.0$) and (b) \textit{$x_2$-axis}: at $x_1 = 0.0$, $x_2 \approx 1.0$.

\subsubsection*{Nonlinearly separable classes}

We construct the \textit{Nonlinear Classifier Neural CRN} (NLC-NCRN) circuit for modeling binary classification tasks with nonlinearly separable classes. We construct this circuit using $f_\theta^{\mathrm{nlcls}}$ as the state dynamics function:
\begin{gather}
    f_{\theta, \beta}^{\mathrm{nlcls}} (\bs x, \bs z) = \theta \bs x + \beta - \bs z\odot\bs z\odot \bs z \label{eq:ftheta_nlcls}
\end{gather} At the end of the feedforward phase, the circuit's output approximates the output of a three-layer FFNN with the \textit{cube root} activation function in its hidden layer: $z_i(T) \rightarrow \sqrt[3]{(\theta \bs x)_i + \beta}$. We discuss the convergence properties and the gradient smoothness of $f_{\theta,\beta}^{\mathrm{nlcls}}$ in SI Text S7. Similar to the nonlinear regression task, this circuit is also ``augmented'' to incorporate the required nonlinearity. Here, we demonstrate the performance of this circuit on two nonlinear classification datasets: Rings2D and XOR2D. Additional details on these simulations are provided in SI Text S8.

\begin{figure}[htbp]
    \centering
    \includegraphics[width=\textwidth]{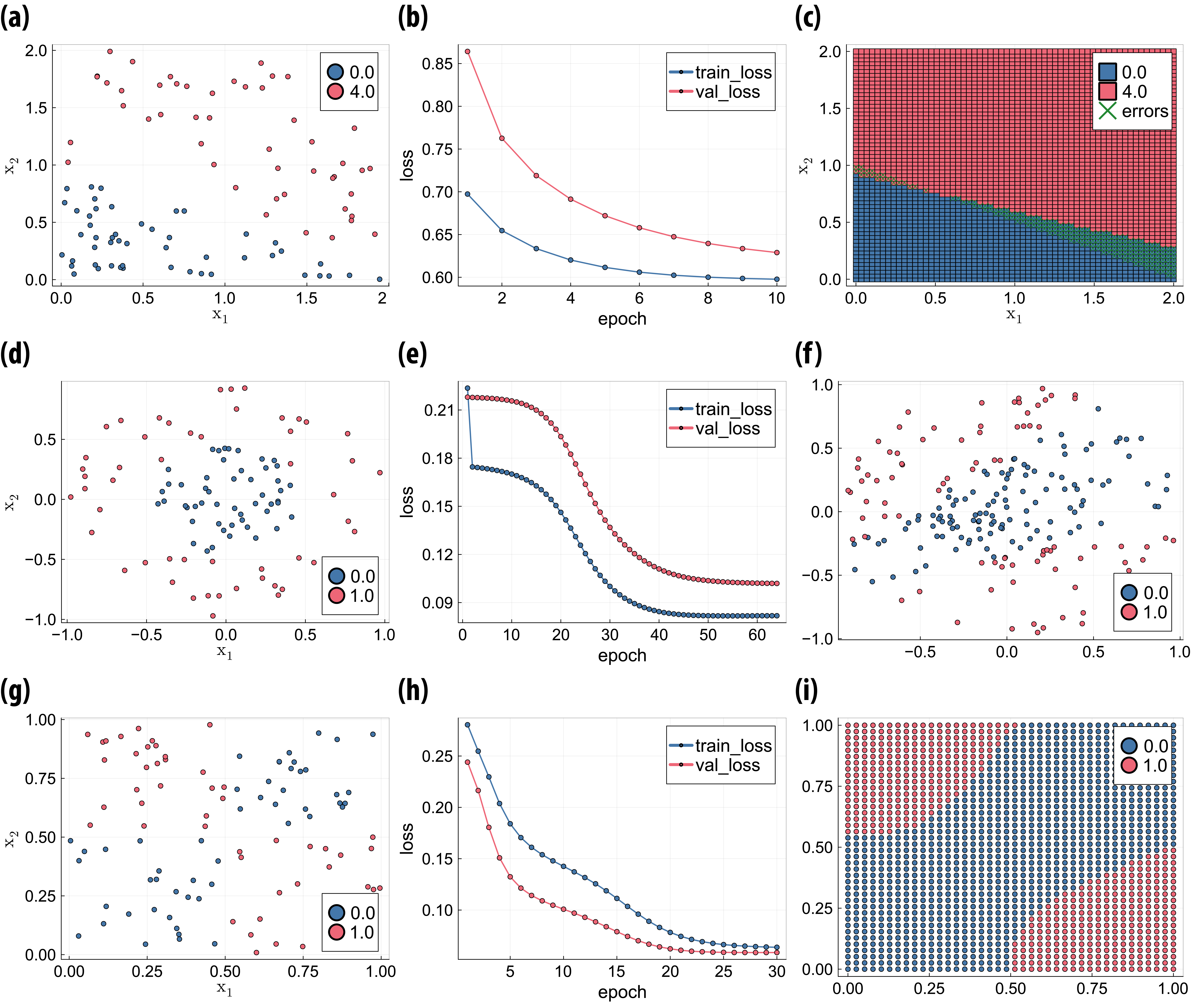}
    \caption{Results of training and inference of the Neural CRN classifiers on (a, b, c) Linear2D, (d, e, f) Rings2D, and (g, h, i) XOR tasks. The first column represents the training datasets of each task. The second column represents the loss curves on training and validation sets showing loss convergence on all three tasks. The third column shows the decision boundaries formed by the trained model on the test set. }
    \label{fig:cls_results}
\end{figure}

\textbf{Datasets.} \textit{Rings2D} is a binary classification dataset with nonlinearly separable classes arranged in concentric rings centered at the origin. The inputs $\bs x\in \mb R^2$ are sampled from the uniform distribution $\bs x \sim \mathrm{Uniform}(-1, 1)$  (Figure~\ref{fig:cls_results}d), and the class labels are assigned based on their Euclidean distance from the origin using three radius hyperparameters $r_1, r_2, r_3 \in \mb R^+$ ($r_1 < r_2 < r_3$) as follows:
\begin{gather*}
    y = \begin{cases}
        y^{\mathrm{OFF}} &  \text{if } 0 < ||\bs x||_2 < r_1 \\
        y^{\mathrm{ON}} & \text{if } r_2 < ||\bs x||_2 < r_3
    \end{cases}.
\end{gather*} The classification parameters are set as follows: $r_1 = 0.45$, $r_2 = 0.5$, and $r_3 = 1$, $y^{\mathrm{ON}} = 1$, $y^{\mathrm{OFF}} = 0$, $\phi = 0.5$.

Figure~\ref{fig:cls_results}d shows the training set, highlighting the concentric ring structure. Figure~\ref{fig:cls_results}e illustrates the loss convergence on the training and validation sets. Figure~\ref{fig:cls_results}f presents model predictions on the test set, with misclassified examples marked by an $\times$. These results confirm that the learned model effectively captures the nonlinear separation boundary between the inner and outer rings.

\textit{XOR2D} is another two-dimensional binary classification dataset with nonlinearly separable classes arranged across quadrants in a unit square anchored at the origin. Geometrically, the \mrm{ON} class occupies the second and fourth quadrants, whereas the \mrm{OFF} class occupies the first and third quadrants of the unit square. Algebraically, these labels are generated by first binarizing the input components $x_1$ and $x_2$ \eqref{eq:booleanize}, and then applying a Boolean \verb*|XOR| function on them \eqref{eq:boolean_label}.

\begin{align}
    x_i^b &= \begin{cases}
        y^{\mathrm{ON}}, \quad \text{if } x_i > \phi^b \\
        y^{\mathrm{OFF}}, \quad \text{otherwise}
    \end{cases}\label{eq:booleanize}\\
    y &= \texttt{XOR}(x_1^b, x_2^b). \label{eq:boolean_label}
\end{align}  The binarization threshold and the classification parameters are set as follows: $\phi^b = 0.5$, $y^{\mathrm{ON}} = 1$, $y^{\mathrm{OFF}} = 0$, and $\phi = 0.5$.

Figure~\ref{fig:cls_results}g shows the training dataset, illustrating the spatial arrangement of the classes in diagonal quadrants. Figure~\ref{fig:cls_results}h shows the loss convergence on both training and validation datasets. Figure~\ref{fig:cls_results}i qualitatively visualizes the separation boundaries learned by the model using a uniformly spaced test grid. Here, we observe that the resulting boundary closely approximates the expected \mrm{XOR}-like boundary, demonstrating the approximation capabilities of the NLC-NCRN circuit. Additional binary nonlinear classification results involving \mrm{AND} and \mrm{OR}-like decision boundaries are demonstrated in SI Text S8.

\textbf{Optimizing the NLC-NCRN circuit.} The cubic polynomial $\fnlcls$ leads to trimolecular reactions in the circuit, posing an implementation challenge. Typically, in practice, trimolecular reactions are decomposed into cascaded bimolecular reactions, leading to larger and slower circuits. To our knowledge, the only other nonlinear analog neural network by Anderson \textit{et al.} \cite{anderson2021reaction} also involves trimolecular reactions. Here, we aim to design an NLC-NCRN circuit that solely uses unimolecular and bimolecular reactions. To this end, we propose a modified dynamics function $f_\theta^{\mathrm{nlclsV2}}$ that supplants the cubic term in $\fnlcls$ with a quadratic term:
\begin{gather}
    f_\theta^{\mathrm{nlclsV2}}  = \theta \bs x - \alpha \bs z \odot \bs z
    \label{eq:ftheta_nlclsV2}
\end{gather} The hyperparameter $\alpha = 0.3$, chosen by trial and error, acts as a damping factor to control the autocatalytic growth of $\bs z$ during the feedback phase. We demonstrate the classification performance of this modified NLC-NCRN circuit in SI Text S9. In future work, we aim to develop a more theoretically grounded framework to inform the choice of $f_\theta$ functions.

\section{Discussion}\label{sec: discussion}

In this work, we introduced a general-purpose chemical neural network (CheNN) architecture, termed Neural CRNs, designed to be scalable and efficient in synthetic biomolecular chemistries. The merits of this framework are primarily due to its use of CRN concentration dynamics to represent neural computations. This analog formulation--aligned with the intrinsic analog nature of chemical kinetics--provides a key advantage over prior implementations and supports further design optimizations. 

The hidden state dynamics function $f_\theta$ is an important design parameter in circuit construction, as it determines the circuit dynamics in both the feed-forward and feedback phases. It should therefore be carefully selected to reflect the complexity of the task at hand. Furthermore, care should be taken to ensure that its dynamics do not diverge within the circuit's runtime.  Although a rigorous framework for selecting an $f_\theta$ does not yet exist, Anderson \textit{ et al.}\cite{anderson2021reaction} laid down a mathematical framework listing a set of properties that an $f_\theta$ must satisfy. We incorporate these principles into our $f_\theta$ design. Finally, for the circuit to be practical, the selected $f_\theta$ should be a lower-order polynomial so that all reactions will be bimolecular or of lower molecularity. In this work, we demonstrated the construction of nonlinear regression and classification models using bimolecular and unimolecular reactions, a significant improvement over prior approaches, which required trimolecular reactions for constructing \mrm{ReLU}-\cite{anderson2021reaction} and \mrm{Tanh}-activated\cite{Arredondo2022-qj} CheNNs. 

Other tunable hyperparameters in the framework include the dimensionality $d$ and the runtime of the analog stages $T$. In most of our demonstrations, we ensured that the hidden state has the same dimensionality as the input (so as to trivialize $W_{\mathrm{in}}$). In cases where they do not match, we may consider this input projection to be a preprocessing step independent of Neural CRN execution, so as to preserve the two-phase execution. The parameter $T$ loosely represents the depth of the network, analogous to the number of hidden layers in conventional neural networks. In the presence of nonlinearity, it additionally represents the extent to which the nonlinearity is applied in the input-to-output transformation. Therefore, this parameter should be tuned in conjunction with $f_\theta$, while being cognizant of the latter's growth rate. A lower value $T$ could lead to underexpression of the nonlinearity, whereas a higher value $T$ could lead to vanishing or exploding gradients \cite{hochreiter1997long} (possibly due to the autocatalytic elements in $f_\theta$). In addition, the values of $y^{\mathrm{ON/OFF}}$ in the classification tasks and the learning rate $\eta$ can also cause gradient instabilities and should be appropriately tuned. Another hyperparameter of note is $\epsilon$, the runtime of the discrete stages. Ideally, its value should be an order of magnitude less than the value of $T$ and should be set based on the relative rates of fast unimolecular and slow bimolecular reactions. 

Supervised learning is an iterative procedure that involves multiple rounds of training and updating parameters with different inputs. Implementing this process in a one-pot reaction model requires additional control modules to reliably transition from one stage to another. One such module that can orchestrate this transition is the \textit{clock mechanism}, where an oscillatory circuit runs in parallel with the main circuit, and the dominant signal in each clock phase conditionally activates the reaction pathways specific to that phase, either by catalysis \cite{vasic2020crn++,Jiang2011-eh} or passive activation \cite{hjelmfelt1991chemical}. However, constructing oscillator circuits in practical synthetic chemistries is challenging because their autocatalytic feedback loops exponentially amplify any noise in the system, destabilizing the circuit rapidly. In this work, we partially mitigate this challenge by streamlining the supervised learning procedure to require only two clock phases (the minimum required). Arredondo \textit{et al.}\cite{Arredondo2022-qj} suggested an alternative clock mechanism employing a microfluidic device that releases clock signals into the circuit at periodic intervals. An asynchronous CheNN implementation such as Chemical Boltzmann Machines \cite{poole2017chemical} does not involve a clock mechanism but relies on impractically high-molecularity reactions, whereas a sequential implementation such as the weight-perturbation algorithm \cite{jabri1992weight} in Arredondo \textit{et al.} \cite{Arredondo2022-qj} requires a large number of clock phases. Our Neural CRN framework strikes an efficient middle-ground, executing the entire supervised learning pipeline within only two clock phases while relying exclusively on unimolecular and bimolecular reactions.

The analog nature of the Neural CRNs framework allows for further circuit optimizations. Observe that $Z$ and $A$ are the only species common between the \mrm{N2} and \mrm{N3} phases. Recall that the $Z$ species are already decoupled between these stages through the creation of proxy $Z^b$ species during the \mrm{N2} stage. By similarly creating proxy $A^b$ species during the \mrm{N2} stage, the two stage pairs \mrm{\{N1, N2\}} and \mrm{\{N3, N4\}} could be executed independently. Although this in itself is not sufficient to eschew the clock mechanism, designing $f_\theta$ so that the resultant CRNs are steady-state systems might allow for their asynchronous execution. In a similar vein, the \mrm{N4} stage alone could be completely detached from the circuit so that gradients accumulate over several iterations. These accumulated gradients can be used to update parameters in an ad hoc manner analogous to batch gradient descent in conventional machine learning. \cite{hinton1992neural} Furthermore, such a decoupled strategy could allow for a wider range of learning rates (currently constrained to $\eta =1$). A more chemically aligned learning scheme could be to train the circuit on the same input until loss convergence before advancing to the next, an approach aligned with some online learning paradigms. We will explore these possibilities in future work.



How does the analog implementation of Neural CRN circuits compare with their discrete CheNN counterparts? We draw comparisons between linear and nonlinear modeling circuits. For linear modeling, we compare our LR-NCRN with the Analog Asymmetric Signal Perceptron (AASP) circuit developed by Banda \textit{et al.} \cite{banda2014learning}, though the comparison is fair only in the case of positive-valued functions (due to the difference in the way they encode negative values). Both circuits are comparable in size (17 species and 18 reactions in AASP vs 17 species and 14 reactions in LR-NCRN). However, AASP is designed as a steady-state system and also places stricter constraints on the reaction rate constants. In contrast, our LR-NCRN is an analog system with uniform rate constants on all reactions, providing a flexible design. For nonlinear modeling, we compare our NLR-NCRN with prior models that implement a three-layer nonlinear feedforward network \cite{Vasic2022-an,Arredondo2022-qj}. The presence of a nonlinearity in the hidden layer of these circuits requires the input-weight integration to be completed before the application of the nonlinear activation. This break in computational flow necessitates the presence of auxiliary control modules that significantly increase the implementation complexity. Furthermore, the discrete nature of these circuits also poses an issue for gradient computation. Our NLR-NCRN, on the other hand, executes nonlinear transformation in the feedforward stage and gradient computation during the feedback phase within single clock cycles. While the use of implicit lifting does increase the circuit size, it could be partially mitigated by sparsifying the parameter matrix $\theta_{\mathrm{sparse}}$, a static version of the \textit{dropout} mechanism  \cite{srivastava2014dropout} used while training machine learning models. 

A potential practical implementation of our Neural CRN system might be possible through DNA-based molecular computing systems \cite{yurke2000dna,lv2021biocomputing, zhang2011dynamic}. In particular, \textit{DNA strand displacement}  (DSD)\cite{qian2011scaling,soloveichik2010dna,srinivas2017enzyme} has emerged as a versatile reaction motif to encode arbitrary chemical dynamics within synthetic biomolecular systems.  In this paradigm, abstract CRNs, which serve as a form of chemical programming language, are systematically compiled into DSD circuits using existing translation schemes. The choice of the translation scheme is guided by the structure and dynamics of the target CRN reactions. Since most reactions in the Neural CRNs framework are non-competitive in nature, suitable translation schemes include the enzyme-free ``two-domain'' strand displacement framework \cite{cardelli2013two} and the enzymatic strand-displacing polymerase-based strand-displacement (PSD) framework \cite{shah2019implementing,baccouche2014dynamic}. In addition, fast annihilative reactions can be implemented using cooperative hybridization reactions in the enzyme-free framework. \cite{zhang2011cooperative} Further, the inherently cascaded nature of multi-reactant DSD circuits naturally facilitates the necessary time-scale separation between unimolecular and multimolecular reactions. Finally, the dynamics of the circuit can be modulated both by engineering the strand displacement rates \cite{zhang2011dynamic} and by varying the concentrations of the fuel complexes. \cite{soloveichik2010dna,Reynaldo2000-kv,shah2019implementing}

However, significant challenges remain before we can experimentally realize Neural CRNs using DSD circuits. The primary obstacle lies in implementing a reliable clock mechanism. Chemical clocks typically rely on autocatalytic reactions \cite{Jiang2011-eh, srinivas2017enzyme, fujii2013predator}, which are difficult to implement in DSD systems due to leakage errors that amplify exponentially in autocatalytic circuits \cite{Reynaldo2000-kv, zhang2011dynamic}. Another challenge is crosstalk in large-scale circuits, due to the limited sequence design space of orthogonal oligonucleotides \cite{milenkovic2005design}. Some of these issues could be mitigated, for example, by expanding the nucleotide alphabet \cite{lee2018genetic}, using nucleotide clamps to minimize spontaneous ``breathing''\cite{wang2017design}, or by developing alternative biomolecular hardware \cite{katz2020dna}. The use of dual-rail encoding presents an additional challenge, as it exponentially increases the circuit size. Exploring asymmetric computational representations, such as those used in the AASP circuit \cite{banda2014learning}, could offer more compact implementations of analog computation and learning. Furthermore, it remains unclear how to incorporate clock signal–based conditional activation into the circuit in a nonintrusive manner. One potential strategy is to use clock signals to activate the non-signal fuel complexes, which in turn initiate the strand displacement reactions. However, precise kinetic control in this approach remains a challenge. An alternative approach might involve using cooperative hybridization strategies \cite{zhang2011cooperative}, where the clock signal and the functional signal co-invade a strand complex to initiate a reaction pathway. Recent studies have introduced various ``timer'' strategies \cite{fern2017dna} designed to activate fuel complexes following a set delay. However, engineering them to be dynamic and periodic remains a challenge. 

Future development of Neural CRN systems should continue to leverage their natural synergy with chemical kinetics to explore several key directions:
(a) discovering alternative state dynamics functions to reduce circuit size and implementation complexity;
(b) developing more compact circuits through simplifying assumptions, such as employing first-order gradient approximations;
(c) improving asynchrony in system execution to reduce reliance on clock mechanisms;
(d) extending the architecture to handle multiclass classification tasks (see preliminary results in SI Text S10); and
(e) extending the Neural CRN systems to process temporal information by integrating the circuit with dynamic memory elements such as chemical delay lines \cite{moles2015delay,banda2014analog} or redesigning the circuit as a reservoir computing system \cite{goudarzi2014comparative} to track changes in a biochemical environment.

\section{Methods}

Training and inference simulations were performed in  Julia (\verb|Julia v1.11.3|) on an Apple Silicon \verb|x64| system. The CRN simulations were performed using the \verb|Catalyst.jl| package using \verb|TRBDF()| as differential equation solver. For faster repeats of experiments, we developed a software tool to generate CRNs by specifying the ODEs in their vector form\footnote{\href{https://github.com/rajiv256/NeuralCRNGen/blob/relu_nofinallayer/relu_main.py}{Code for the tool}} The details for the simulation setup of all the demonstrations presented in this work are provided in the SI text.

\section{Conclusion}

In this work, we introduced Neural CRNs, a synthetic chemical learning framework implemented using deterministic CRNs. Unlike prior architectures that chemically mimic the algebra of neural networks, Neural CRNs use chemical reactions directly as input/output devices, resulting in concise and simpler reaction systems. The novelty of our approach lies in unifying the theory of Neural ODEs with CRN theory, and integrating them into a coherent chemical learning framework. In this work, we presented a streamlined supervised learning procedure, separating the discrete and analog computations in the framework into separate stages and enforcing time-scale separation between them, so that the entire learning process can be implemented within two clock phases. We then performed several proof-of-concept demonstrations, including linear and nonlinear regression and classification tasks, to validate the framework and its supervised learning procedure. Notable improvements include: (a) the construction of a minimal-size learning circuit for linear regression comprising 15 species and 13 reactions, (b)  a significant reduction in the circuit size for nonlinear regression through a performance-preserving simplification that involves computing approximate gradients, and (c) a nonlinear classifier circuit composed solely of unimolecular and bimolecular reactions. Finally, we present a plausible road map towards a synthetic biochemical implementation, identify key design and engineering challenges, and propose plausible solutions. Overall, our Neural CRNs framework offers a novel paradigm for building adaptive biochemical circuits, laying a foundation for future applications in synthetic biology, bioengineering, and adaptive biomedicine.

\section{Author Contributions}
John Reif (JR) conceived the problem of online learning and gradient calculation in a chemical medium. Rajiv Nagipogu (RN) proposed the solution of using analog neural networks and designed the supervised learning procedure. RN also developed the simulation experiments, defined the supervised learning tasks, and constructed the corresponding Neural CRN architectures. JR provided guidance and critical feedback throughout the project. RN edited the manuscript, and JR contributed to revisions and corrections.

\section{Conflicts of Interest}
conflicts of interest: no

\section{Funding}
This work was funded by the National Science Foundation under grant nos. 1909848 and 2113941 to JR.

\section{Acknowledgements}
The authors thank the reviewers for their constructive feedback, which helped significantly improve this manuscript.

\section{Supplementary Information}

More details on the supervised learning architecture, including code descriptions, CRNs used, and hyperparameter settings in different model configurations.  Additional simulation results of nonlinear regression and classification using alternative hidden state dynamics functions and additional datasets. Proof-of-concept demonstrations of various simplifying assumptions, such as the minimal linear regression circuit and first-order gradient approximations in nonlinear regression and classification circuits. Proofs of asymptotic stability for state dynamics functions. Evidence of implicit lifting in a nonlinear classification task. Preliminary implementation of multiclass classification. A comparison of gradient and parameter trajectories between a Neural CRN and a reference Neural ODE framework. 

\clearpage

\begin{figure}
    \centering
    \includegraphics[width=\textwidth]{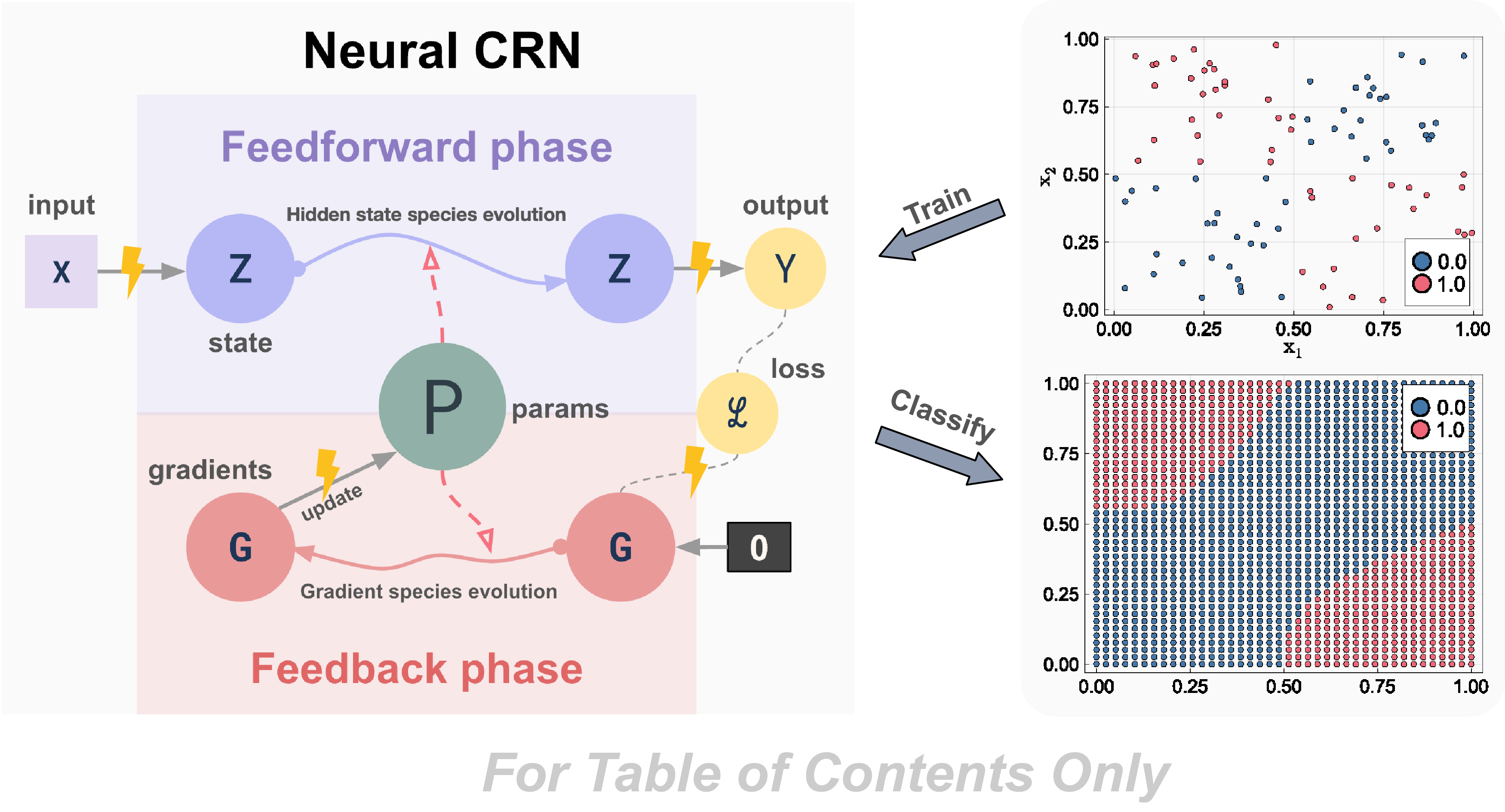}
\end{figure}

\clearpage

\bibliography{references}%

\end{document}